\documentclass[10pt,twocolumn,letterpaper]{article}

\usepackage{iccv}
\usepackage{times}
\usepackage{epsfig}
\usepackage{graphicx}
\usepackage{amsmath}
\usepackage{amssymb}

\usepackage{graphicx}
\usepackage{wrapfig}

\usepackage[table]{xcolor}
\usepackage{color}
  \definecolor{c0}{HTML}{663300}
 \definecolor{c1}{HTML}{c91f37}
 \definecolor{c1_2}{HTML}{336699}
 \definecolor{c2}{HTML}{FF9900}
 \definecolor{c3}{HTML}{4b5cc4}
 \definecolor{c4}{HTML}{0eb83a}
  \definecolor{c5}{HTML}{ef7a82}
  \definecolor{c6}{HTML}{9933FF}
  
    \definecolor{c7}{HTML}{33CC33}

    \usepackage{booktabs}
\usepackage{multirow}

\usepackage{amsmath,amssymb} 
\usepackage[table]{xcolor}
\usepackage{array}
\usepackage{dsfont}

\usepackage[pagebackref=true,breaklinks=true,letterpaper=true,colorlinks,bookmarks=false]{hyperref}

\iccvfinalcopy 

\def\iccvPaperID{xxxx} 

\begin{document}

\title{Learning Occupancy for Monocular 3D Object Detection}


\author{
Liang Peng$^{1, 3}$ 
\quad Junkai Xu$^{1, 3}$ 
\quad Haoran Cheng${^{1, 3}}$ 
\quad Zheng Yang${^{3}}$ 
\quad Xiaopei Wu${^{1}}$ \\ 
\quad Wei Qian${^{3}}$ 
\quad Wenxiao Wang ${^{2}}$ 
\quad Boxi Wu${^{2}}$ 
\quad Deng Cai $^1$
\quad \\
\textsuperscript{\rm 1}State Key Lab of CAD\&CG, Zhejiang University \\
\textsuperscript{\rm 2}School of Software Technology, Zhejiang University \\
\textsuperscript{\rm 3}FABU Inc. \\
{\tt\small  \{pengliang, xujunkai, haorancheng\}@zju.edu.cn}
}

\maketitle

\begin{abstract}
   Monocular 3D detection is a challenging task due to the lack of accurate 3D information.
   Existing approaches typically rely on geometry constraints and dense depth estimates to facilitate the learning, but often fail to fully exploit the benefits of three-dimensional feature extraction in frustum and 3D space.
   In this paper, we propose \textbf{OccupancyM3D}, a method of learning occupancy for monocular 3D detection.
   It directly learns occupancy in frustum and 3D space, leading to more discriminative and informative 3D features and representations.
   Specifically, by using synchronized raw sparse LiDAR point clouds, we define the space status and generate voxel-based occupancy labels.
   We formulate occupancy prediction as a simple classification problem and design associated occupancy losses.
   Resulting occupancy estimates are employed to enhance original frustum/3D features.
   As a result, experiments on KITTI and Waymo open datasets demonstrate that the proposed method achieves a new state of the art and surpasses other methods by a significant margin.
   Codes and pre-trained models will be available at \url{https://github.com/SPengLiang/OccupancyM3D}.
\end{abstract}

\section{Introduction}
	Three dimensional object detection is a critical task in many real-world applications, such as self-driving and robot navigation.
	Early methods \cite{qi2018frustum,zhou2018voxelnet,shi2020pv} typically rely on LiDAR sensors because they can produce sparse yet accurate 3D point measurements. 	
	In contrast, cameras provide dense texture features but lack 3D information. 
	Recently, monocular-based methods \cite{Monodle,CaDDN,GUPNet,peng2022did} for 3D detection, also known as monocular 3D detection, have gained significant attention from both industry and academia due to their cost-effectiveness and deployment-friendly nature.

	Recovering accurate 3D information from a single RGB image poses a challenge.
	While previous researches have employed geometry constraints \cite{Deep3DBox,RTM3D,MonoRUn,GUPNet} and dense depth estimates \cite{PseudoLidar,PatchNet,CaDDN} to facilitate 3D reasoning, they often overlook the importance of discriminative and informative 3D features in 3D space, which are essential for effective 3D detection. 
	They mainly focus on improving features in 2D space, with little attention paid to better feature encoding and representation in the frustum and 3D space.

	\begin{figure}[tbp]
\centering 
		\includegraphics[width=1.0\linewidth]{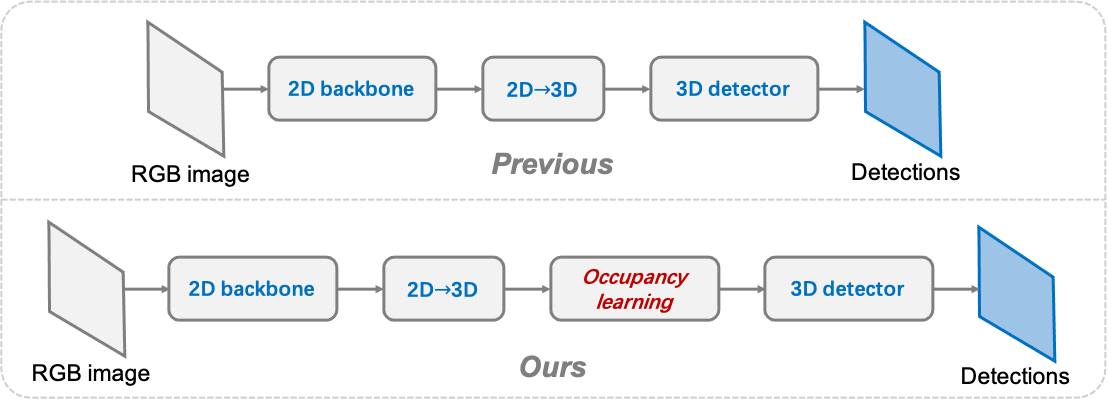}      
		\caption{
				Overall design.
				We introduce occupancy learning for monocular 3D detection.
				Best viewed in color.
				}
		\label{fig:intro}
		\vspace{-2mm}
\end{figure}
	
	Towards this goal, in this paper we propose to learn occupancy in frustum and 3D space, to obtain more discriminative and informative 3D features/representations for monocular 3D detection.
	Specifically, we employ synchronized raw sparse LiDAR point clouds to generate voxel-based occupancy labels in frustum and 3D space during the training stage.
	Concerning the sparsity of LiDAR points, we define three occupancy statuses: \textit{free, occupied, and unknown}.
	Based on this, we voxelize the 3D space and use ray tracing on each LiDAR point to obtain occupancy labels.
	With the occupancy labels, we can enforce explicit 3D supervision on intermediate 3D features.
	It allows the network to learn voxelized occupancy for current 3D space, which enhances the original 3D features.
	This process is also performed in the frustum space, enabling a more fine-grained manner in extracting three-dimensional features for near objects due to the perspective nature of camera imagery.
	Overall, we call the proposed occupancy learning method \textbf{OccupancyM3D}, and illustrate the framework overview in Figure \ref{fig:intro}.
	
	To demonstrate the effectiveness of our method, we conduct experiments on the competitive KITTI and Waymo open datasets.
	As a result, the proposed method achieves state-of-the-art results with a significant margin over other methods.
	Our contributions can be summarized as follows:
		
\begin{itemize}
		\item 
		We emphasize the importance of feature encoding and representation in the frustum and 3D space for monocular 3D detection, and we propose to learn occupancy in both space.
		\item 
		We propose a method to generate occupancy labels using synchronized raw sparse LiDAR points and introduce corresponding occupancy losses, enabling the network to learn voxelized occupancy in both frustum and 3D space. 
		This occupancy learning process facilitates the extraction of discriminative and informative 3D features in the network.
		\item 
		Experiments demonstrate the superiority of the proposed method. 
		Evaluated on challenging KITTI and Waymo open datasets, our method achieves new state-of-the-art (SOTA) results and outperforms other methods by a significant margin.
\end{itemize}

\section{Related Work}
\subsection{LiDAR Based 3D Object Detection}
	LiDAR-based methods \cite{wu2022transformation,PART,lang2019pointpillars,yang2019std,SA-SSD,zheng2021se,SFD} currently dominate 3D object detection accuracy because of their precise depth measurements. 
	Due to the unordered nature of point clouds, LiDAR-based methods are required to organize the input data. 
	There are four main streams based on the input data representation: point-based, voxel-based, range-view-based, and hybrid-based. 
	PointNet families \cite{PointNet,PointNet++} are effective methods for feature extraction from raw point clouds, allowing point-based methods \cite{qi2018frustum,Pointrcnn,Point-gnn,yang20203dssd} to directly perform 3D detection.
	 Voxel-based methods \cite{zhou2018voxelnet,yan2018second,Yin_2021_CVPR,zheng2021se,VoxelRCNN} organize point clouds into voxel grids, making them compatible with regular convolutional neural networks.
	 Range-view-based methods \cite{bewley2020range,fan2021rangedet,chai2021point} convert point clouds into range-view to accommodate the LiDAR scanning mode. 
	 Hybrid-based methods \cite{PVRCNN,yang2019std,chen2020every,noh2021hvpr}  use a combination of different representations to leverage their individual strengths. 
	 There is still a significant performance gap between monocular and LiDAR-based methods, which encourages researchers to advance monocular 3D detection.


\subsection{Monocular 3D Object Detection}

	Significant progress has been made in advancing monocular 3D detection in recent years.
	The ill-posed problem of recovering instance level 3D information from a single image is challenging and important, attracting many researches.
	This is also the core sub-problem in monocular 3D detection.
	Early works \cite{Mono3D,Deep3DBox} resort to using scene priors and geometric projections to resolve objects' 3D locations.
	More recent monocular methods \cite{M3D,RoI10D,RTM3D,AutoShape,MonoFlex,li2022diversity,MonoJSG}  employ more geometry constraints and extra priors like CAD models to achieve this goal. 
	AutoShape \cite{AutoShape} incorporates shape-aware 2D/3D constraints into the 3D detection framework by learning distinguished 2D and 3D keypoints.
	MonoJSG \cite{MonoJSG} reformulates the instance depth estimation as a progressive refinement problem and propose a joint semantic and geometric cost volume to model the depth error.
	As RGB images lack explicit depth information, many works rely on dense depth estimates.
	Some methods \cite{PseudoLidar,AM3D,PatchNet} directly convert depth map to pseudo LiDAR or 3D coordinate patches, and some works  \cite{CaDDN} use depth distributions to lift 2D image features to 3D space.
	Therefore, previous well-designed LiDAR 3D detectors can be easily employed on such explicit 3D features.
	Other researches \cite{D4LCN,DDMP-3D,DD3D,chong2022monodistill,chen2022pseudo,peng2022did} also take advantage of depth maps or LiDAR point clouds as guidance for feature extraction and auxiliary information. 
	While previous works have leveraged geometry constraints and dense depth estimates, they have not fully explored feature encoding and representation in the frustum and 3D space. 
	To address this, our proposed method focuses on learning occupancy for monocular 3D detection.

\subsection{3D Scene Representations}
	Recent researches \cite{occupancyNet,nerf,nerf++} rapidly advance implicit representations.
	Implicit representations have the advantage of arbitrary-resolution on modeling the 3D scene.
	This nature is beneficial for fine-grained tasks such as 3D reconstruction and semantic segmentation.
	Different from them, monocular 3D detection is an instance level task, and we explore explicit occupancy learning using fixed-sized voxels.
	Implicit occupancy representations for this task can be explored in future works, which is an interesting and promising topic.
	Additionally, many bird's-eye-view (BEV) based works \cite{OFTNet,CaDDN,BEVDet,PETR,BEVFormer,BEVDepth} have been proposed recently.
	These works commonly employ BEV representations and obtain great success, especially for multi-camera BEV detection.
	The most related work to ours is CaDDN \cite{CaDDN}.
	We follow its architecture design except for the proposed occupancy learning module, and we replace its 2D backbone with lightweight DLA34 \cite{DLA34}.
	It should be noted that our work focuses on the monocular setting, and extending the method to the multi-camera setup is a potential avenue for future researches.

\begin{figure*}[htbp]
\centering 
				\vspace{-5mm}  
		\includegraphics[width=1.03\linewidth]{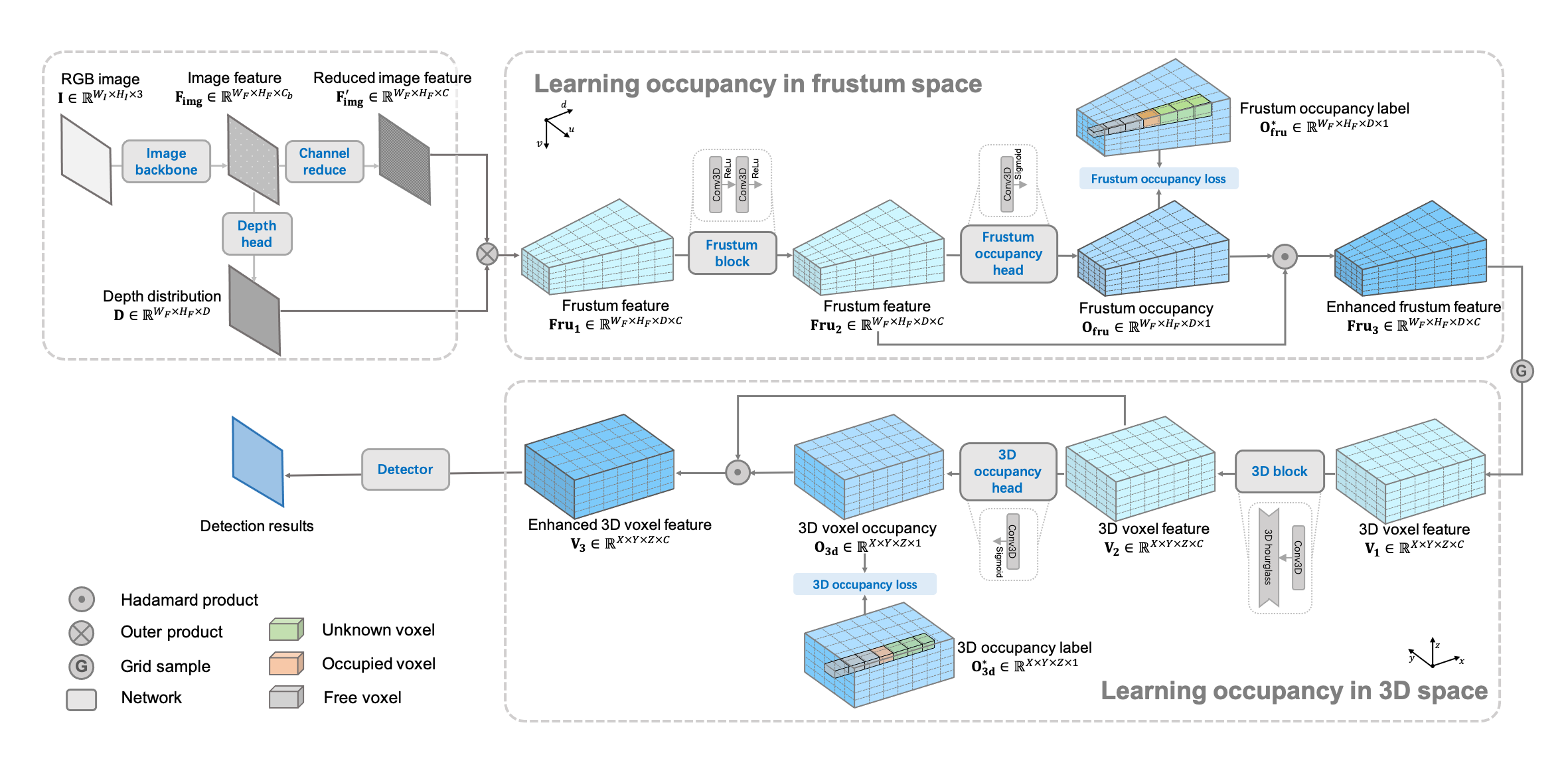}       
				\vspace{-5mm}  
		\caption{
				Network overview.
				Compared to previous works, our method employs two newly-proposed components for learning occupancy in frustum and 3D space.
				All network blocks in the proposed parts consist of vanilla 3D convolutions.
				Please refer to Section \ref{sec:net} for detailed feature passing description.
				For occupancy in frustum and 3D space and their network blocks, please see Section \ref{sec:space};
				For occupancy label generation, please see Section \ref{sec:label};
				For occupancy losses, please see Section \ref{sec:occ_loss};
				Best viewed in color with zoom in.
				}
		\label{fig:overview}
\end{figure*}

\section{OccupancyM3D}

\subsection{Preliminary and Overview} 
\noindent
	{\bf{Task Definition.}} 
	We first describe the preliminary of this task and the method.
	At inference, monocular 3D detection takes only a single RGB image and outputs interested amodal 3D bounding boxes in the current scene.
	At the training stage, our method requires RGB images, 3D box labels annotated on LiDAR points and synchronized LiDAR data.
	It is worth noting that the system has been calibrated, and the camera intrinsics and extrinsics between the camera and LiDAR are available.

\noindent
	{\bf{Network Overview.} {\label{sec:net}}}
	We present the network overview of our method in Figure \ref{fig:overview}.
	First, a single RGB image is fed into the DLA34 \cite{DLA34} backbone network to extract features.
	Then, we use these features to produce categorical depth distributions \cite{CaDDN}, which lifts 2D features to frustum space.
	After that, we employ the depth predictions and backbone features to generate frustum features.
	They are used for {\bf{\textit{learning occupancy in frustum space}}}, and then are transformed to voxelized 3D features using grid-sampling.
	Such voxelized 3D features are employed to {\bf{\textit{study occupancy in 3D space}}}. 
	Occupancy learning in both frustum and 3D spaces can produce reasonable occupancy estimates that enhance the original features.
	The final enhanced voxelized 3D features are passed to the detection module to obtain final 3D detection results.
	
	At the training stage, occupancy estimates are supervised by the generated occupancy labels in frustum and 3D space, respectively, using the proposed occupancy losses.
	We detail the occupancy learning in following sections.

\subsection{Occupancy Learning}
	We consider a frustum voxel or regular 3D voxel to be occupied if it contains part of an object. 
	We denote the resulting voxel states as \textbf{frustum occupancy} and \textbf{3D occupancy}, respectively.

	In this section, we introduce occupancy learning for monocular 3D detection.
	It is organized as four parts: {\bf\textit{occupancy in frustum/3D space}}, {\bf\textit{occupancy labels}}, {\bf\textit{occupancy losses}}, and {\bf\textit{occupancy and depth}}.

\subsubsection{Occupancy in Frustum Space and 3D Space}{\label{sec:space}}
	After extracting backbone features, we employ a depth head to obtain dense category depth \cite{CaDDN}.
	To save GPU memory, we use a convolution layer to reduce the number of feature channels, and the resulting feature is lifted to a frustum feature $\mathbf{Fru_1} \in  \mathbb{R}^{W_F\times H_F\times D \times C}$ with the assistance of depth estimates.
	Then we extract frustum feature $\mathbf{Fru_2} \in  \mathbb{R}^{W_F\times H_F\times D \times C}$ as follows:
\begin{equation}
\mathbf{Fru_2} = f_1(\mathbf{Fru_1})
\label{equ:fru_2}
\end{equation}
	where $f_1$ denotes two 3D convolutions followed by ReLU activate functions.
	Then we use a 3D convolution layer $f_2$ and sigmoid function to obtain frustum occupancy $\mathbf{O_{fru}}\in  \mathbb{R}^{W_F\times H_F\times D\times1}$, which is supervised by corresponding labels $\bf O^*_{fru}$ as described in Section \ref{sec:label} and Section \ref{sec:occ_loss}.
\begin{equation}
\mathbf{O_{fru}} = {\rm Sigmoid}(f_2(\mathbf{Fru_2}))
\label{equ:fru_2}
\end{equation}
	The frustum occupancy indicates the feature density in the frustum space, thus inherently can be employed to weight original frustum features  for achieving enhanced frustum feature $\mathbf{Fru_3} \in  \mathbb{R}^{W_F\times H_F\times D\times C}$ as follows:
\begin{equation}
\mathbf{Fru_3} = \mathbf{O_{fru}} \odot \mathbf{Fru_2}
\label{equ:Fru_3}
\end{equation}
	where $\odot$ denotes the Hadamard product (element-wise multiplication).
	The resulted frustum feature $\bf Fru_3$ is transformed to regular voxelized feature $\mathbf{V_1}  \in  \mathbb{R}^{X\times Y\times Z\times C}$ via grid-sampling \cite{CaDDN}.
	The occupancy learning process is then repeated in the regular 3D space. 
\begin{equation}
\mathbf{V_2} = f_3(\mathbf{V_1}); \mathbf{O_{3d}} = {\rm Sigmoid}(f_4(\mathbf{V_2})); \mathbf{V_3} = \mathbf{O_{3d}} \odot \mathbf{V_2}
\label{equ:V_2}
\end{equation}
	To better encode 3D features in the regular 3D space, we use a 3D hourglass-like design \cite{PSMNet} in $f_3$, and $f_4$ is a 3D convolution.
	Finally, we have more informative 3D voxel feature $\mathbf{V_3}\in  \mathbb{R}^{X\times Y\times Z\times C}$ for the detection module.
	Please refer to the supplementary material for more detailed network architecture and their ablations.
	
\noindent
	{\textbf{\textit{What is the rationale behind learning occupancy in both frustum and 3D space?}}}
	Occupancy learning in both frustum and 3D space is beneficial  because they have different nature.
	Frustum space has a resolution that depends on camera intrinsics and the downsample factor of the backbone network, while voxelized 3D space has a resolution that is decided by the pre-defined voxel size and detection range.
	Frustum voxels are irregular and vary in size based on the distance to the camera, which results in fine-grained voxels for objects that are closer and coarse-grained voxels for objects that are far away. In contrast, regular 3D voxels have the same size throughout the 3D space.	
	On the other hand, frustum space is more fit to camera imagery, but objects in the frustum space cannot precisely represent the real 3D geometry.
	Thus the feature extraction and occupancy in frustum space have distortion for objects/scenes.
	Therefore, occupancy learning in both frustum and 3D space is complementary, and can result in more informative representations and features.

\subsubsection{Occupancy Labels}
\begin{figure}[t]
\centering 
		\includegraphics[width=0.9\linewidth]{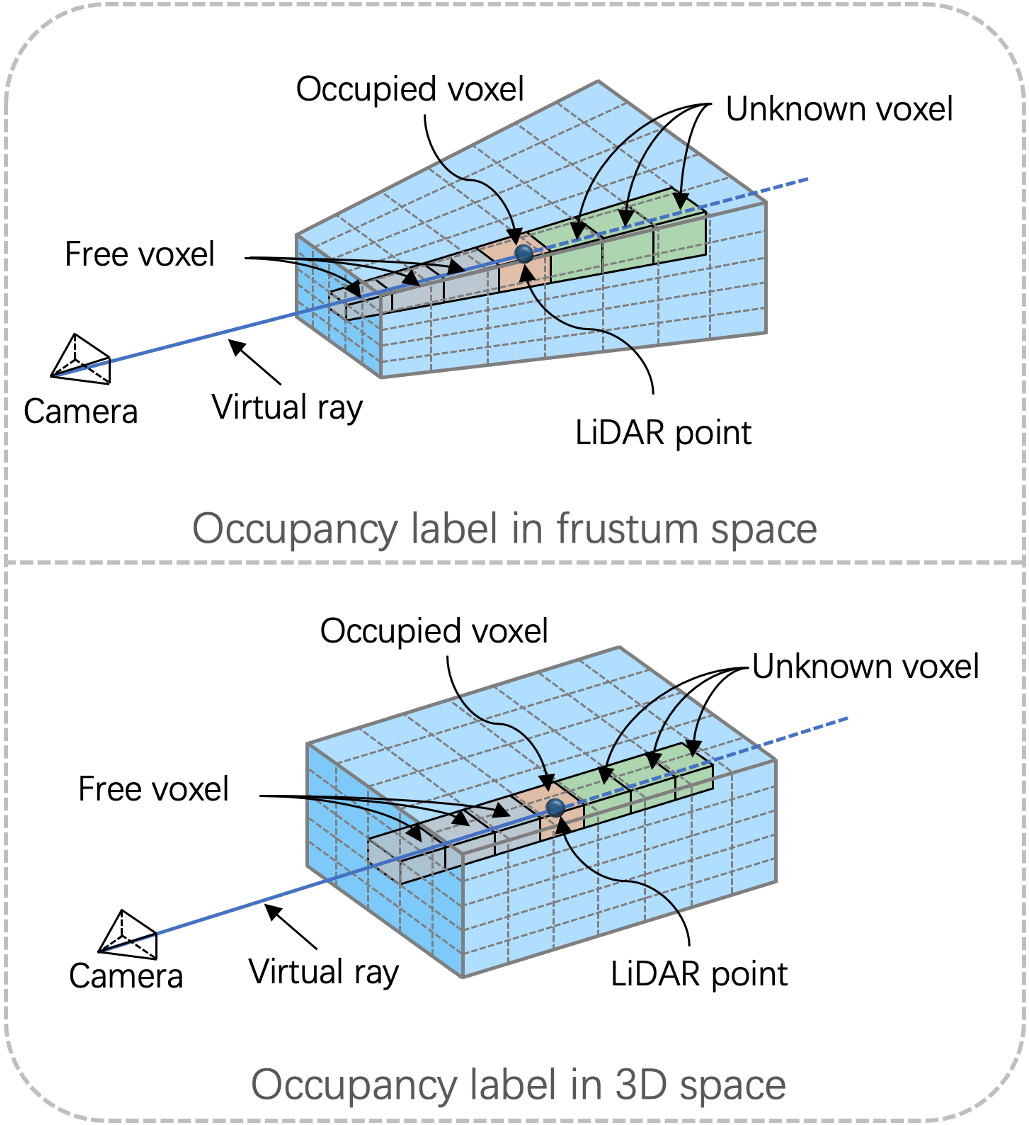}      
		\caption{
				Occupancy label generation in frustum and 3D space.
				Best viewed in color with zoom in.
				}
		\label{fig:label}
\end{figure}

{\label{sec:label}}
	Given a set of sparse LiDAR points $\mathbf{P} \in \mathbb{R}^{N\times 3}$, where $N$ is the number of points and $3$ is the coordinate dimension ($X, Y, Z$), we generate corresponding occupancy labels. 
	The process is illustrated in Figure \ref{fig:label}, and is operated on every LiDAR point.
	More formally, we first define three space status and represent them with numbers: \textbf{\textit{free:0}}, \textbf{\textit{occupied:1}}, \textbf{\textit{unknown:-1}}.
	We then describe the occupancy label generation process in the frustum and 3D space, respectively.

	{\textbf{Occupancy label in frustum space:}} 
	Let us denote the frustum occupancy label as $\mathbf{O^*_{fru}}\in \mathbb{R}^{W_F\times H_F\times D}$, where $W_F$ and $H_F$ are feature resolution, and $D$ is the depth category. 
	We first project LiDAR points onto the image plane to form a category depth index map. 
	Each valid projected point has a category depth index, while invalid points (no LiDAR points projections) are given negative indexes of $-1$.
	This index map is then downsampled to fit the feature resolution, resulting in $\mathbf{Ind} \in \mathbb{R}^{W_F\times H_F}$.
	Benefit from the camera projection nature, we can easily distinguish the space status as follows:
	\begin{equation}
\mathbf{O^*_{fru}}_{i,j,d}=\left\{
\begin{array}{ccl}
\vspace {0.3cm}
1 \quad  &      & {if \hspace {0.1cm}  \mathbf{Ind}_{i,j} > -1 \hspace {0.1cm}  and \hspace {0.1cm}   d = \mathbf{Ind}_{i,j},}\\
\vspace {0.3cm}
0 \quad  &      & {if \hspace {0.1cm}  \mathbf{Ind}_{i,j} > -1  \hspace {0.1cm}  and \hspace {0.1cm}  d < \mathbf{Ind}_{i,j},}\\
-1    &      & {otherwise}.\\ 
\end{array} \right. 
\label{equ:label_fru}
\end{equation}
	where $i,j,d\in W_F, H_F, D$. 
	Note that we do not consider unknown voxels in both the occupancy labels and occupancy losses. 
	We use the known voxels, \textit{i.e.}, the free and occupied voxels, to perform occupancy learning.

{\textbf{Occupancy label in 3D space:}} 
	We denote  $\mathbf{O^*_{3d}}\in  \mathbb{R}^{X\times Y\times Z}$ as the 3D occupancy label, where $X,Y,Z$ are determined by the pre-defined voxel size and detection range.
	We voxelize LiDAR points within the grid and set the voxels containing points to $1$, and those without points to $-1$.
	In this way, occupied voxels can be easily achieved.
	To obtain the free voxels,  we utilize ray tracing from each LiDAR point to the camera, where intersected voxels are set as free, filled by 0.
	We summarize the occupancy label in 3D space as follows: 
\begin{equation}
\mathbf{O^*_{3d}}_{x,y,z}=\left\{
\begin{array}{ccl}
\vspace {0.3cm}
1 \quad  &      & {if \hspace {0.1cm} \mathbf{Vol_{3d}}_{x,y,z} > 0,}\\
\vspace {0.3cm}
0 \quad  &      & {if \hspace {0.1cm}   {\rm{R}}(\mathbf{O^*_{3d}}_{x,y,z}) \cap Ray_{point\rightarrow cam}}\\
-1    &      & {otherwise}.\\ 
\end{array} \right. 
\label{equ:label_3d}
\end{equation}
	where $x,y,z\in X, Y, Z$.
	 In this equation, $\mathbf{Vol_{3d}}$ denotes the voxelized grid. 
	 $\mathbf{Vol_{3d}}>0$ when it is occupied by LiDAR points.
	${\rm R}(\cdot)$ refers to the voxel range and  ${\rm{R}}(\mathbf{O_{3d}}_{i,j,d}) \cap Ray_{point\rightarrow cam}$ denotes that the voxel at index $i,j,d$ intersects with a ray from LiDAR points to the camera.
	In this way, 3D occupancy labels are generated.

	When generating voxel-based occupancy labels, there is a quantization error that arises due to the discretization process. 
	A smaller voxel size results in lower quantization error, providing more fine-grained and accurate information.
	However, it requires more computation and GPU memory resources.

\subsubsection{Occupancy Losses}{\label{sec:occ_loss}}
	We use generated occupancy labels $\mathbf{O^*_{fru}}$ and $\mathbf{O^*_{3d}}$ to supervised the predicted occupancy  $\mathbf{O_{fru}}$ and $\mathbf{O_{3d}}$, respectively.
	We regard occupancy prediction as a simple classification problem and use focal loss \cite{focal} as the classification loss. 
	Only valid voxels, \textit{i.e.}, free and occupied voxels, contribute to the loss, and unknown voxels are ignored. 
	We first obtain valid masks $\mathbf{M_{fru}}\in  \mathbb{R}^{W_F\times H_F\times D}$ and $\mathbf{M_{3d}} \in  \mathbb{R}^{X\times Y\times Z}$.
	$\mathbf{M_{fru}} = true$ if  $\mathbf{O^*_{fru}}>-1$ otherwise $false$.
	$\mathbf{M_{3d}}$ is obtained using the similar way.
	
	Therefore, the occupancy loss in frustum space is:
\begin{equation}
\mathcal{L}_{fru}={\rm{FL}}(\mathbf{O^*_{fru}}[\mathbf{M_{fru}}], \mathbf{O_{fru}}[\mathbf{M_{fru}}])
\label{equ:loss_fru}
\end{equation}
	where $\rm FL(\cdot)$ refers to focal loss.
	Similarly, we can obtain 3D occupancy loss as follows:
\begin{equation}
\mathcal{L}_{3d}={\rm{FL}}(\mathbf{O^*_{3d}}[\mathbf{M_{3d}}], \mathbf{O_{3d}}[\mathbf{M_{3d}}])
\label{equ:loss_3d}
\end{equation}
	The final occupancy loss is their sum:
\begin{equation}
\mathcal{L}_{occupancy}=\mathcal{L}_{fru} + \mathcal{L}_{3d}
\label{equ:loss_occ}
\end{equation}
	The occupancy loss allows the network to learn informative and discriminative features and representations, thus benefit downstream tasks.
	Therefore, the final loss of the network is:
\begin{equation}
\mathcal{L}=\mathcal{L}_{org} +  \lambda \mathcal{L}_{occupancy}
\label{equ:loss}
\end{equation}
	where $\mathcal{L}_{org}$ denotes the original detection and depth losses in CaDDN \cite{CaDDN} and $\lambda$ is the occupancy loss weighting factor, which is set to $1$ by default.

\subsubsection{Occupancy and Depth}
	Occupancy has some similarities with 2D depth map, especially the frustum occupancy.
	They both can represent object geometry surface in the space.
	However, depth map is two-dimensional while occupancy is three-dimensional.
	Occupancy is beyond the depth and can base on it.
	It is able to express dense features of objects but not only the surface.
	For unknown space due to occlusion, the occupancy can infer reasonable results.
	Moreover, learning occupancy in frustum and 3D space allows the network to study more informative features under a higher dimension compared to 2D space.
	
	Occupancy and depth are not mutually exclusive representations.
	In fact, they complement each other in the 3D object detection task.
	Without depth, the network has to deal with a large search space, making it challenging to learn reasonable occupancy features.
	Incorporating depth estimation provides the network with a good starting point and facilitates learning the occupancy features. 
	Therefore, it is recommended to utilize both depth and occupancy information to achieve better representations and features for monocular 3D detection.

		  \begin{table*}[h]
	  \small
      \centering
						\begin{tabular}{l|c|c|ccc|ccc}
				\toprule   
				\multirow{2}{*}{Approaches} &\multirow{2}{*}{Venue} &\multirow{2}{*}{Input} & \multicolumn{3}{c|}{AP$_{BEV}$ (IoU=0.7)$|\scriptstyle R_{40}$} & \multicolumn{3}{c}{AP$_{3D}$ (IoU=0.7)$|\scriptstyle R_{40}$} \\ 
				~&~& ~ & Easy & Moderate & Hard& Easy & Moderate & Hard\\ 
				\midrule         
				Kinematic3D \cite{Kinematic3D}& \textit{ECCV20}& Video  & 26.69 & 17.52 & 13.10  & 19.07 & 12.72 & 9.17 \\
				DfM \cite{DFM}& \textit{ECCV22}& Video  & 31.71 & \bf \textcolor{blue}{22.89} & \bf \textcolor{blue}{19.97}  & 22.94 & \bf \textcolor{blue}{16.82} & \bf \textcolor{blue}{14.65} \\
				\midrule
				AM3D \cite{AM3D}& \textit{ICCV19}& Image & 25.03 & 17.32 & 14.91 & 16.50 & 10.74 & 9.52 \\
				M3D-RPN \cite{M3D} & \textit{ICCV19}& Image& 21.02  & 13.67  & 10.23  & 14.76  & 9.71  & 7.42 \\
				MonoPair \cite{MonoPair}& \textit{CVPR20}& Image & 19.28 & 14.83 & 12.89 & 13.04 & 9.99 & 8.65\\
				D4LCN \cite{D4LCN}& \textit{CVPR20}& Image & 22.51 & 16.02 & 12.55 & 16.65 & 11.72 & 9.51 \\
				PatchNet \cite{PatchNet}& \textit{ECCV20}& Image& 22.97 & 16.86 & 14.97 & 15.68 & 11.12 & 10.17\\
				RTM3D \cite{RTM3D} & \textit{ECCV20}& Image& 19.17  & 14.20  &  11.99   & 14.41  & 10.34  &  8.77   \\ 
				Ground-Aware \cite{Ground-Aware}& \textit{RAL21}& Image& 29.81& 17.98 & 13.08 & 21.65& 13.25 & 9.91  \\
				Monodle \cite{Monodle} & \textit{CVPR21}& Image & 24.79 & 18.89 & 16.00  & 17.23 & 12.26 & 10.29\\
				DDMP-3D \cite{DDMP-3D}& \textit{CVPR21}& Image  & 28.08 & 17.89 & 13.44  & 19.71 & 12.78 & 9.80\\
				GrooMeD-NMS \cite{GrooMeD-NMS}& \textit{CVPR21}& Image& 26.19 & 18.27 & 14.05 & 18.10 & 12.32 & 9.65\\
				MonoRUn \cite{MonoRUn}& \textit{CVPR21}& Image& 27.94 & 17.34 & 15.24  &19.65 & 12.30 & 10.58\\
				MonoEF \cite{MonoEF}& \textit{CVPR21}& Image & 29.03 & 19.70 & 17.26   & 21.29 & 13.87 & 11.71\\
				MonoFlex \cite{MonoFlex}& \textit{CVPR21}& Image   & 28.23 & 19.75 & 16.89 & 19.94 & 13.89 & 12.07  \\
				CaDDN \cite{CaDDN} & \textit{CVPR21}& Image & 27.94 & 18.91 & 17.19  & 19.17 & 13.41 & 11.46 \\
				MonoRCNN \cite{MonoRCNN}& \textit{ICCV21}& Image &  25.48 & 18.11 & 14.10  & 18.36 & 12.65 & 10.03\\
				GUP Net \cite{GUPNet}  & \textit{ICCV21}& Image  & 30.29 & 21.19 & 18.20 & 22.26 & 15.02 & 13.12  \\
				AutoShape \cite{AutoShape}& \textit{ICCV21}& Image &  30.66 & 20.08 & 15.95  & 22.47 & 14.17 & 11.36\\
				PCT \cite{PCT} & \textit{NeurIPS21}& Image  & 29.65 & 19.03 & 15.92 & 21.00 & 13.37  & 11.31\\
				MonoCon \cite{MonoCon} & \textit{AAAI22}& Image  & 31.12 & 22.10 & 19.00 &  22.50 & 16.46 &13.95  \\
				HomoLoss \cite{HomoLoss}  & \textit{CVPR22}& Image  & 29.60 & 20.68 & 17.81 & 21.75 & 14.94 & 13.07 \\
				MonoDTR \cite{MonoDTR}  & \textit{CVPR22}& Image  & 28.59 & 20.38 & 17.14 & 21.99 & 15.39 & 12.73  \\
				MonoJSG \cite{MonoJSG}  & \textit{CVPR22}& Image  & 32.59 & 21.26 & 18.18 & \bf \textcolor{blue}{24.69} & 16.14 & 13.64  \\
				DEVIANT \cite{kumar2022deviant}  & \textit{ECCV22}& Image  &29.65 &20.44 &17.43 & 21.88& 14.46 &11.89  \\
				DID-M3D \cite{peng2022did}& \textit{ECCV22}& Image   & \bf \textcolor{blue}{32.95} & 22.76 & 19.83  & 24.40 & 16.29 & 13.75 \\
				\midrule 
				{\bf 	OccupancyM3D}&\multirow{1}{*}{-} &  Image&\bf \textcolor{red}{35.38} &\bf \textcolor{red}{24.18}  &\bf \textcolor{red}{21.37}  &\bf \textcolor{red}{25.55}  & \bf \textcolor{red}{17.02}  & \bf \textcolor{red}{14.79} \\
				\bottomrule 
			\end{tabular}
				  			\caption{
			Comparisons on KITTI \textit{test} set for \textit{Car} category. 
			The \textcolor{red}{red} refers to the highest result and \textcolor{blue}{blue} is the second-highest result.
			Our method outperforms other methods including monocular and video-based methods.
			}
			\label{tab:test}
  \end{table*}

\begin{table*}[t]
    \setlength\tabcolsep{2pt}
    \small
    \centering
    \begin{tabular}{l|c|cccc|cccc}
    \toprule
    \multirow{3}*{Methods} &\multirow{2}{*}{Venue}  & \multicolumn{4}{c|}{3D mAP / mAPH (IoU = 0.7)} & \multicolumn{4}{c}{3D mAP / mAPH (IoU = 0.5)} \\
   ~&~ & Overall &0 - 30m & 30 - 50m & 50m - $\infty$ &Overall &0 - 30m & 30 - 50m & 50m - $\infty$\\
    \midrule
    \multicolumn{10}{c}{\textit{Comparison on LEVEL 1}} \\
    \midrule
    PatchNet\cite{PatchNet} &\textit{ECCV20}  &0.39/0.37   &1.67/1.63   &0.13/0.12   &0.03/0.03   &2.92/2.74   &10.03/9.75  &1.09/0.96   &0.23/0.18 \\
    CaDDN \cite{CaDDN}   &\textit{CVPR21}   &\textbf{\textcolor{blue}{5.03/4.99}}   &\textbf{\textcolor{blue}{14.54/14.43}}   &\textbf{\textcolor{blue}{1.47/1.45}}   &\textbf{\textcolor{blue}{0.10/0.10}} &17.54/17.31   &\textbf{\textcolor{blue}{45.00/44.46}}   &9.24/9.11   &0.64/0.62 \\
    PCT \cite{PCT} &\textit{NeurIPS21}   &0.89/0.88   &3.18/3.15   &0.27/0.27   &0.07/0.07   &4.20/4.15   &14.70/14.54   &1.78/1.75   &0.39/0.39 \\
    MonoJSG \cite{MonoJSG}  &\textit{CVPR22}   &0.97/0.95   &4.65/4.59   &0.55/0.53   &0.10/0.09   &5.65/5.47   &20.86/20.26   &3.91/3.79   &0.97/0.92 \\
    DEVIANT  \cite{kumar2022deviant}   &\textit{ECCV22}  &2.69/2.67   &6.95/6.90   &0.99/0.98   &0.02/0.02   &10.98/10.89   &26.85/26.64   &5.13/5.08   &0.18/0.18  \\
    DID-M3D \cite{peng2022did} &\textit{ECCV22}   & -/-   & -/-   & -/-   & -/-   & \textbf{\textcolor{blue}{20.66/20.47}}   &40.92/40.60   &\textbf{\textcolor{blue}{15.63/15.48}}   &\textbf{\textcolor{red}{5.35/5.24}}\\
  \midrule
     \bf OccupancyM3D  &\textit{-}  & \textbf{\textcolor{red}{10.61/10.53}} &\textbf{\textcolor{red}{29.18/28.96}} & \textbf{\textcolor{red}{4.49/4.46}} & \textbf{\textcolor{red}{0.41/0.40}} & \textbf{\textcolor{red}{28.99/28.66}} &\textbf{\textcolor{red}{61.24/60.63}} & \textbf{\textcolor{red}{23.25/23.00}} &\textbf{\textcolor{blue}{3.65/3.59}} \\
    \midrule
   \multicolumn{10}{c}{\textit{Comparison on LEVEL 2}} \\
    \midrule
        PatchNet\cite{PatchNet} &\textit{ECCV20}   & 0.38/0.36   &1.67/1.63   &0.13/0.11   &0.03/0.03   &2.42/2.28   &10.01/9.73  &1.07/0.94   &0.22/0.16\\
    CaDDN \cite{CaDDN}   &\textit{CVPR21}   &\textbf{\textcolor{blue}{4.49/4.45}}   &\textbf{\textcolor{blue}{14.50/14.38}}   &\textbf{\textcolor{blue}{1.42/1.41}}   &\textbf{\textcolor{blue}{0.09/0.09}}   &16.51/16.28   &\textbf{\textcolor{blue}{44.87/44.33}}   &8.99/8.86   &0.58/0.55\\
    PCT  \cite{PCT} &\textit{NeurIPS21}  &0.66/0.66   &3.18/3.15   &0.27/0.26   &0.07/0.07   &4.03/3.99   &14.67/14.51   &1.74/1.71   &0.36/0.35\\
    MonoJSG  \cite{MonoJSG}  &\textit{CVPR22}  &0.91/0.89   &4.64/4.65   &0.55/0.53   &0.09/0.09   &5.34/5.17   &20.79/20.19   &3.79/3.67   &0.85/0.82\\
    DEVIANT  \cite{kumar2022deviant}  &\textit{ECCV22}   &2.52/2.50   &6.93/6.87   &0.95/0.94   &0.02/0.02   &10.29/10.20   &26.75/26.54   &4.95/4.90   &0.16/0.16\\
    DID-M3D\cite{peng2022did}  &\textit{ECCV22}  & -/-   & -/-   & -/-   & -/-   &\textbf{\textcolor{blue}{19.37/19.19}}   &40.77/40.46   &\textbf{\textcolor{blue}{15.18/15.04}}  &\textbf{\textcolor{red}{4.69/4.59}}\\
    \midrule
        {\bf 	OccupancyM3D}  &\textit{-}   & \textbf{\textcolor{red}{10.02/9.94}} &\textbf{\textcolor{red}{28.38/28.17}} & \textbf{\textcolor{red}{4.38/4.34}} & \textbf{\textcolor{red}{0.36/0.36}} & \textbf{\textcolor{red}{27.21/26.90}} &\textbf{\textcolor{red}{61.09/60.49}} & \textbf{\textcolor{red}{22.59/22.34}} & \textbf{\textcolor{blue}{3.18/3.13}}\\
    \bottomrule
    \end{tabular}
    \caption{
    		Results on WaymoOD \emph{val} set for \textit{Vehicle} category. 
    		The \textcolor{red}{red} refers to the highest result and \textcolor{blue}{blue} is the second-highest result.
		Our method outperforms other methods by significant margins on most metrics.
		Note that our method has the detection range limitation of $[2, 59.6](meters)$, while the perspective-view based method DID-M3D \cite{peng2022did}  does not have this shortcoming.
		Thus our method performs worse for objects within $[50m, \infty]$ under IoU=0.5 criterion.
		}
    \label{tab:waymo}
\end{table*}

  		  \begin{table*}[h]
	  \small
      \centering
						\begin{tabular}{l|c|c|ccc|ccc}
				\toprule   
				\multirow{2}{*}{Approaches} &\multirow{2}{*}{Venue} &\multirow{2}{*}{Input} & \multicolumn{3}{c|}{Pedestrian  AP$_{BEV}$/AP$_{3D}$(IoU=0.5)$|\scriptstyle R_{40}$} & \multicolumn{3}{c}{Cyclist  AP$_{BEV}$/AP$_{3D}$(IoU=0.5)$|\scriptstyle R_{40}$} \\ 
				~&~& ~ & Easy & Moderate & Hard& Easy & Moderate & Hard\\ 
				\midrule
				DfM \cite{DFM}& \textit{ECCV22}& Video  & -/13.70 & -/8.71 & -/7.32 &  -/\bf \textcolor{red}{8.98} &-/\bf \textcolor{red}{5.75} &-/\bf \textcolor{red}{4.88}  \\
				\midrule
				Monodle \cite{Monodle} & \textit{CVPR21}& Image & 10.73/9.64 & 6.96/6.55 & 6.20/5.44  & 5.34/4.59 & 3.28/2.66 & 2.83/2.45\\
				DDMP-3D \cite{DDMP-3D}& \textit{CVPR21}& Image  & 5.53/4.93 & 4.02/3.55 & 3.36/3.01  & 4.92/4.18 & 3.14/2.50 & 2.44/2.32\\
				MonoRUn \cite{MonoRUn}& \textit{CVPR21}& Image& 11.70/10.88 & 7.59/6.78 & 6.34/5.83  &1.14/1.01 & 0.73/0.61 & 0.66/0.48\\
				MonoEF \cite{MonoEF}& \textit{CVPR21}& Image & 4.61/4.27 & 3.05/2.79 & 2.85/2.21  & 2.36/1.80 & 1.18/0.92 & 1.15/0.71\\
				MonoFlex \cite{MonoFlex}& \textit{CVPR21}& Image   & 10.36/9.43 & 7.36/6.31 & 6.29/5.26 & 4.41/4.17 & 2.67/2.35 & 2.50/2.04  \\
				CaDDN \cite{CaDDN} & \textit{CVPR21}& Image & 14.72/12.87 & 9.41/8.14 & 8.17/6.76  & {\bf \textcolor{red}{9.67}}/7.00 & {\bf \textcolor{red}{5.38}}/3.41 & {\bf \textcolor{red}{4.75}}/\bf \textcolor{blue}{3.30} \\
				GUP Net \cite{GUPNet}  & \textit{ICCV21}& Image  & {\bf \textcolor{blue}{15.62}}/\bf \textcolor{red}{14.95} & {\bf \textcolor{blue}{10.37}}/\bf \textcolor{red}{9.76} & {\bf \textcolor{blue}{8.79}}/\bf \textcolor{red}{8.41} & 6.94/5.58 & 3.85/3.21 & 3.64/2.66  \\	
				AutoShape \cite{AutoShape}& \textit{ICCV21}& Image &  -/5.46 & -/3.74 & -/3.03 &  -/5.99 &-/3.06 & -/2.70 \\
				MonoCon \cite{MonoCon} & \textit{AAAI22}& Image  & -/13.10 & -/8.41 & -/6.94 &  -/2.80 &-/1.92 & -/1.55  \\   
				HomoLoss \cite{HomoLoss}  & \textit{CVPR22}& Image  & 13.26/11.87 & 8.81/7.66 & 7.41/6.82 & 6.81/5.48 & 4.09/3.50 & 3.78/2.99 \\		
				MonoJSG \cite{MonoJSG}  & \textit{CVPR22}& Image  &  -/11.02 & -/7.49 & -/6.41 &  -/5.45 &-/3.21 &-/2.57  \\
				DEVIANT \cite{kumar2022deviant}  & \textit{ECCV22}& Image & 14.49/13.43 & 9.77/8.65 & 8.28/7.69 &  6.42/5.05 & 3.97/3.13 &3.51/2.59 \\
				\midrule 
				{\bf 	OccupancyM3D}&\multirow{1}{*}{-} &  Image&\bf \textcolor{red}{16.54}/\bf \textcolor{blue}{14.68} &\bf \textcolor{red}{10.65}/\bf \textcolor{blue}{9.15}  &\bf \textcolor{red}{9.16}/\bf \textcolor{blue}{7.80}  &\bf \textcolor{blue}{8.58}/\bf \textcolor{blue}{7.37}  & \bf \textcolor{blue}{4.35}/\bf \textcolor{blue}{3.56}  & {\bf \textcolor{blue}{3.55}}/2.84 \\
				\bottomrule 
			\end{tabular}
				  			\caption{
			Comparisons on KITTI \textit{test} set for \textit{Pedestrian} and \textit{Cyclist} categories. 
			The \textcolor{red}{red} refers to the highest result and \textcolor{blue}{blue} is the second-highest result.
			Our method achieves new state-of-the-art results. 
			}
			\label{tab:test2}
  \end{table*}

\section{Experiments}

\subsection{Implementation Details}
	We employ PyTorch \cite{Pytorch} for implementation.
	The network is trained on 4 NVIDIA 3080Ti (12G) GPUs, with a total batch size of 8 for 80 epochs.
	We use Adam \cite{Adam} optimizer with initial learning rate 0.001 and employ the one-cycle learning rate policy \cite{One}. 
	We use pre-trained DLA34 \cite{DLA34} backbone from \cite{DD3D}.
	We employ flip and crop data augmentation \cite{GUPNet}.
	For KITTI \cite{KITTI2012}, we fix the input image to $1280 \times 384$, detection range $[2, 46.8] \times [-30.08, 30.08] \times  [-3.0, 1.0] (meter)$  for $x,y,z$ axes under the LiDAR coordinate system, respectively.
	We use voxel size $[0.16, 0.16, 0.16] (meter)$.
	For Waymo  \cite{waymo}, we downsample the input RGB image from $1920 \times 1280$ to $960 \times 640$ to meet GPU memory. 
	We use detection range $[2, 59.6] \times [-25.6, 25.6] \times  [-2.0, 2.0] (meter)$  for $x,y,z$ axes due to the larger depth domain on Waymo.
	We use voxel size $[0.16, 0.16, 0.16] (meter)$.
	\textbf{\textit{Due to the space limitation, more network details and experimental results are provided in the supplementary material}}.

\subsection{Datasets and Metrics}
	Following the fashion in previous works, we conduct experiments on competitive KITTI and Waymo open datasets.
	
\noindent	
	{\textbf{KITTI:}} KITTI \cite{KITTI2012} is a widely employed benchmark for autonomous driving.
	KITTI3D object dataset consists of 7,481 training samples and 7,518 testing samples, where labels on \textit{test} set keep secret and the final performance is evaluated on KITTI official website.
	To conduct ablations, the training samples are further divided into a \textit{train} set and a \textit{val} set \cite{3DOP}.
	They individually contain 3,512 and 3,769 samples, respectively. 
	KITTI has three categories: \textit{Car}, \textit{Pedestrian}, and \textit{Cyclist}.
	According to difficulties (2D box height, occlusion and truncation levels), KITTI divides objects into \textit{Easy}, \textit{Moderate}, and \textit{Hard}.	
	Following common practice \cite{AP40,CaDDN,AutoShape}, we use AP$_{BEV}|\scriptstyle R_{40}$ and AP$_{3D}|\scriptstyle{R_{40}}$ under $IoU$  threshold of  $0.7$ to evaluate the performance.
	
\noindent	
{\textbf{Waymo:}}
	Waymo open dataset (WaymoOD) \cite{waymo} is a large modern dataset for autonomous driving.
	It has 798 sequences for training and 202 sequences for validation.
	Following previous works \cite{CaDDN,peng2022did}, we use the front camera of the multi-camera rig and provide performance comparison on \textit{val} set for the vehicle category.
	To make fair comparisons, we use one third samples of training sequences to train the network due to the large-scale and high frame rate of this dataset.
	Waymo divides objects to LEVEL 1 and LEVEL 2 according to the LiDAR point number within objects.
	For metrics, we employ the official mAP and mAPH  under LEVEL 1 and LEVEL 2.

		\begin{figure*}[h]
\centering 
		\includegraphics[width=1.0\linewidth]{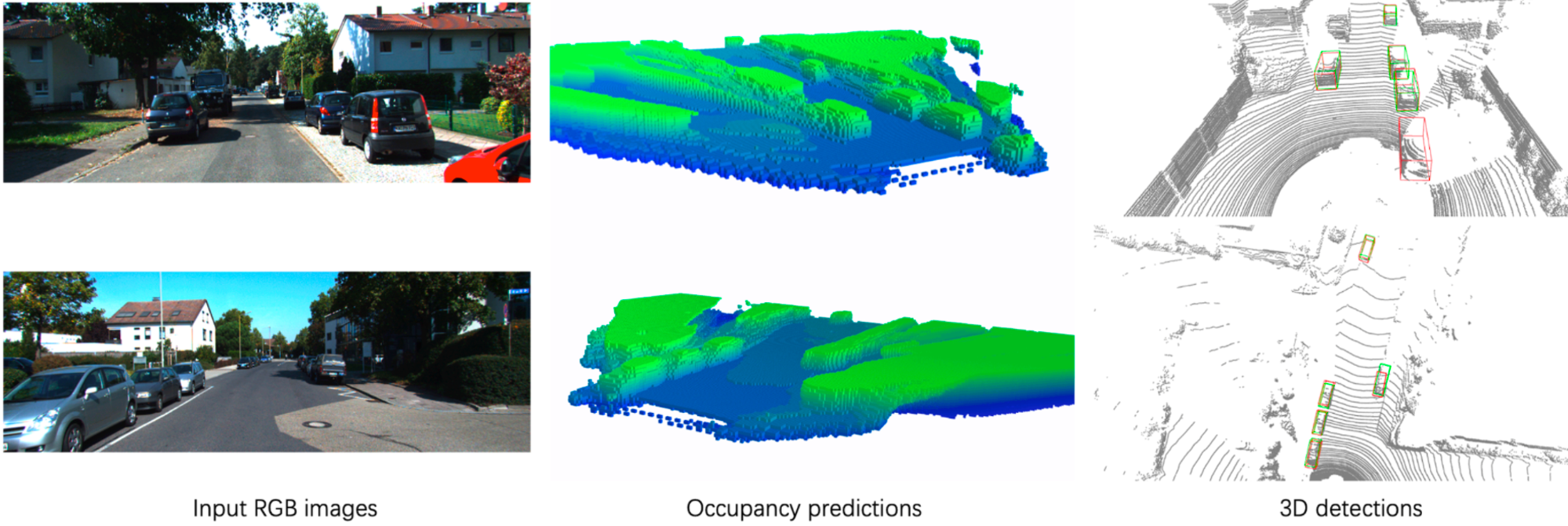}      
		\caption{
				Qualitative results of occupancy predictions and 3D detections on KITTI \textit{val} set.
				In 3D detections, {\color{red}{Red}} boxes are our results and  {\color{green}{Green}} boxes denote ground-truths.
				The LiDAR point clouds in 3D detections are used only for visualization.
				We can see that the proposed method generates reasonable occupancy predictions for the current scene, which benefits downstream monocular 3D detection task.
				However, our method may fail to estimate heavily occluded objects (see right objects of the bottom picture).
				More qualitative results and discussions are provided in the supplementary material. 
				Best viewed in color with zoom in.
				}
		\label{fig:qual}
\end{figure*}

\subsection{Results on KITTI and Waymo Datasets}
	We provide the performance comparisons on KITTI and WaymoOD.
	Table \ref{tab:test} shows the results of \textit{Car} category  on KITTI \textit{test} set. 
	Our method outperforms other methods including video-based methods with a large margin.
	For example, the proposed method exceeds CaDDN \cite{CaDDN} under all metrics, \textit{e.g.}, 25.55/17.02/14.79 \textit{vs.} 19.17/13.41/11.46 AP$_{3D}$.
	Our method outperforms DID-M3D \cite{peng2022did} by a margin of 2.43/1.42/1.54 AP$_{BEV}$.
	When compared to the recent video-based method DfM \cite{DFM}, OccupancyM3D also shows better performance, \textit{e.g.}, 24.18 \textit{vs.}  22.89 AP$_{BEV}$ under the moderate setting.
	In Table \ref{tab:test2},  we provide comparisons on other categories, namely, \textit{Pedestrian} and \textit{Cyclist}.
	The results demonstrate the superiority of our method on different categories.
	To sum up, our method achieves new state-of-the-art results on KITTI \textit{test} set for monocular 3D detection.
	
	We also evaluate our method on Waymo open dataset (WaymoOD) \cite{waymo} and obtain promising results.
	As shown in Table \ref{tab:waymo}, our method surpasses other methods with a significant margin.
	For example, under LEVEL 1 setting, OccupancyM3D outperforms CaDDN \cite{CaDDN} by 5.58/5.54 mAP/mAPH (10.61/10.53 \textit{vs.} 5.03/4.99) and 11.55/11.35 mAP/mAPH (28.99/28.66 \textit{vs.} 17.54/17.31)  with IoU 0.7 and 0.5 criterions, respectively.
	Compared to DID-M3D \cite{peng2022did}, under IoU criterion 0.5, our method outperforms it by 8.33/8.19 mAP/mAPH (28.99/28.66 \textit{vs.} 20.66/20.47) and 7.84/7.71 mAP/mAPH (27.21/26.90 \textit{vs.} 19.37/19.19) with LEVEL 1 and 2 settings, respectively.
	This success can be attributed to the fact that occupancy learning benefits from the diverse scenes present in large datasets.
	In other words, large datasets especially favor the proposed occupancy learning method.
	Interestingly, concerning objects within $[50m, \infty]$, our method performs worse than DID-M3D \cite{peng2022did}.
	It is because our method is voxel-based, which has a detection range limitation ($[2, 59.6] (meters)$ in our method).
	By contrast, DID-M3D is a perspective-based method, indicating that it does not have this limitation and can detect more faraway objects.
	We encourage future works to address this range limitation in our method.

\subsection{Ablations}
	Following common practice in previous works, we perform ablations on KITTI \textit{val} set to validate the effectiveness of each component.
	We compare the performance on \textit{Car} category under IoU criterion 0.7.
	Due to the space limitation, we provide only the main ablation in the main text, and \textbf{\textit{more detailed ablations are provided in the supplementary material}}.
	
	We provide the main ablation in Table \ref{tab:main_abala}.
	It can be easily seen that occupancy learning significantly benefit the final detection performance.
	When enforcing occupancy learning in frustum space, the detection AP$_{3D}$ increases from 21.04/17.05/15.01 to 24.69/17.79/15.16 (Exp. (a)$\rightarrow$(b)).
	On the other hand,  when enforcing occupancy learning in 3D space, the detection AP$_{3D}$ is boosted to 24.64/18.88/16.38 (Exp. (c)).
	Finally, the model obtains 5.83/2.91/2.14 AP$_{3D}$ gains (Exp. (a)$\rightarrow$(d)) by employing occupancy learning in both frustum and 3D space.
	This main ablation demonstrates the effectiveness of our method.

\begin{table}[t]
\centering
\footnotesize
\begin{tabular}{c|cc|ccc}
				\toprule  
				\multirow{2}{*}{E.} & \multirow{2}{*}{OL-FS} &\multirow{2}{*}{OL-3DS} &  \multicolumn{3}{c}{AP$_{BEV}$/AP$_{3D}$ (IoU=0.7)$|\scriptstyle R_{40}$} \\ 
				~&~&~ &  Easy & Moderate & Hard \\ 
				\midrule    
				   (a)&~&~&30.32/21.04  & 24.58/17.05 & 22.02/15.01 \\
				    (b)& \checkmark&~&35.46/24.69 & 25.46/17.79 & 22.96/15.16 \\
				   (c)& ~&\checkmark&33.15/24.64 & 25.45/18.88 & 22.68/16.38 \\
				    (d)& \checkmark&\checkmark&\bf 35.72/26.87 &\bf 26.60/19.96 &\bf 23.68/17.15  \\
					\bottomrule 
			\end{tabular}
			        		\caption{ 
		Main ablation. ``E." in the table is the experiment ID; ``OL-FS"  refers to occupancy learning in frustum space; ``OL-3DS" denotes occupancy learning in 3D space.
		}
		\label{tab:main_abala}
\end{table}

\subsection{Qualitative Results}
	We present qualitative results of occupancy predictions and 3D detections in Figure \ref{fig:qual}.
	Our method can predict reasonable occupancy for the current scene, especially for foreground objects.
	This indicates the potential of occupancy learning in downstream tasks.
	Nevertheless, we can see that the occupancy estimates are not very accurate for heavily occluded objects (see right objects of the bottom picture), which leaves room for improvement in future works.
	Concerning the space limitation, more qualitative results are included in the supplementary material, with providing more detailed discussions.

\section{Limitation and Future Work}
	One significant drawback of this work is the voxel size limitation. 
	Large voxels in explicit voxel-based representation can reduce computation overhead and GPU memory, but at the cost of failing to precisely describe the 3D geometry of the scene due to quantization errors. 
	Conversely, smaller voxel sizes are able to express fine-grained 3D geometry but come at the significant expense of increased computation overhead and GPU memory usage. 	
	On the other hand, the voxel-based method has limited detection ranges.
	This work mainly focuses on occupancy learning in the monocular 3D detection task, and the exploration of its application in more downstream tasks such as multi-camera detection and segmentation is less explored.
	We believe that it is an interesting and promising topic and encourage future works to alleviate the above limitations to advance the self-driving community.

\section{Conclusion}
	In this paper, we propose to learn occupancy for monocular 3D detection, to obtain more discriminative and informative 3D features.
	To perform occupancy learning, we design occupancy labels by using synchronized raw sparse LiDAR point clouds and introduce corresponding occupancy losses.
	Ablations verify the effectiveness of each proposed component.
	To the best of our knowledge, this is the first work that introduces occupancy learning to monocular 3D detection.
	We conduct experiments on the challenging KITTI and Waymo open datasets.
	The results demonstrate that the proposed method achieves new state-of-the-art results and outperforms other methods by a large margin.

{\small
\bibliographystyle{ieee_fullname}
\bibliography{egbib}

\begin{thebibliography}{10}\itemsep=-1pt

\bibitem{bewley2020range}
Alex Bewley, Pei Sun, Thomas Mensink, Dragomir Anguelov, and Cristian
  Sminchisescu.
\newblock Range conditioned dilated convolutions for scale invariant 3d object
  detection.
\newblock {\em arXiv preprint arXiv:2005.09927}, 2020.

\bibitem{M3D}
Garrick Brazil and Xiaoming Liu.
\newblock M3d-rpn: Monocular 3d region proposal network for object detection.
\newblock In {\em Proceedings of the IEEE International Conference on Computer
  Vision}, pages 9287--9296, 2019.

\bibitem{Kinematic3D}
Garrick Brazil, Gerard Pons-Moll, Xiaoming Liu, and Bernt Schiele.
\newblock Kinematic 3d object detection in monocular video.
\newblock In {\em European Conference on Computer Vision}, pages 135--152.
  Springer, 2020.

\bibitem{chai2021point}
Yuning Chai, Pei Sun, Jiquan Ngiam, Weiyue Wang, Benjamin Caine, Vijay
  Vasudevan, Xiao Zhang, and Dragomir Anguelov.
\newblock To the point: Efficient 3d object detection in the range image with
  graph convolution kernels.
\newblock In {\em Proceedings of the IEEE/CVF Conference on Computer Vision and
  Pattern Recognition}, pages 16000--16009, 2021.

\bibitem{PSMNet}
Jia-Ren Chang and Yong-Sheng Chen.
\newblock Pyramid stereo matching network.
\newblock In {\em Proceedings of the IEEE Conference on Computer Vision and
  Pattern Recognition}, pages 5410--5418, 2018.

\bibitem{MonoRUn}
Hansheng Chen, Yuyao Huang, Wei Tian, Zhong Gao, and Lu Xiong.
\newblock Monorun: Monocular 3d object detection by reconstruction and
  uncertainty propagation.
\newblock In {\em Proceedings of the IEEE/CVF Conference on Computer Vision and
  Pattern Recognition}, pages 10379--10388, 2021.

\bibitem{chen2020every}
Qi Chen, Lin Sun, Ernest Cheung, and Alan~L Yuille.
\newblock Every view counts: Cross-view consistency in 3d object detection with
  hybrid-cylindrical-spherical voxelization.
\newblock {\em Advances in Neural Information Processing Systems},
  33:21224--21235, 2020.

\bibitem{Mono3D}
Xiaozhi Chen, Kaustav Kundu, Ziyu Zhang, Huimin Ma, Sanja Fidler, and Raquel
  Urtasun.
\newblock Monocular 3d object detection for autonomous driving.
\newblock In {\em Proceedings of the IEEE Conference on Computer Vision and
  Pattern Recognition}, pages 2147--2156, 2016.

\bibitem{3DOP}
Xiaozhi Chen, Kaustav Kundu, Yukun Zhu, Huimin Ma, Sanja Fidler, and Raquel
  Urtasun.
\newblock 3d object proposals using stereo imagery for accurate object class
  detection.
\newblock {\em IEEE transactions on pattern analysis and machine intelligence},
  40(5):1259--1272, 2017.

\bibitem{MonoPair}
Yongjian Chen, Lei Tai, Kai Sun, and Mingyang Li.
\newblock Monopair: Monocular 3d object detection using pairwise spatial
  relationships.
\newblock In {\em Proceedings of the IEEE/CVF Conference on Computer Vision and
  Pattern Recognition}, pages 12093--12102, 2020.

\bibitem{chen2022pseudo}
Yi-Nan Chen, Hang Dai, and Yong Ding.
\newblock Pseudo-stereo for monocular 3d object detection in autonomous
  driving.
\newblock In {\em Proceedings of the IEEE/CVF Conference on Computer Vision and
  Pattern Recognition}, pages 887--897, 2022.

\bibitem{chong2022monodistill}
Zhiyu Chong, Xinzhu Ma, Hong Zhang, Yuxin Yue, Haojie Li, Zhihui Wang, and
  Wanli Ouyang.
\newblock Monodistill: Learning spatial features for monocular 3d object
  detection.
\newblock {\em arXiv preprint arXiv:2201.10830}, 2022.

\bibitem{VoxelRCNN}
Jiajun Deng, Shaoshuai Shi, Peiwei Li, Wengang Zhou, Yanyong Zhang, and
  Houqiang Li.
\newblock Voxel r-cnn: Towards high performance voxel-based 3d object
  detection.
\newblock In {\em Proceedings of the AAAI Conference on Artificial
  Intelligence}, volume~35, pages 1201--1209, 2021.

\bibitem{D4LCN}
Mingyu Ding, Yuqi Huo, Hongwei Yi, Zhe Wang, Jianping Shi, Zhiwu Lu, and Ping
  Luo.
\newblock Learning depth-guided convolutions for monocular 3d object detection.
\newblock In {\em Proceedings of the IEEE/CVF Conference on Computer Vision and
  Pattern Recognition}, pages 11672--11681, 2020.

\bibitem{fan2021rangedet}
Lue Fan, Xuan Xiong, Feng Wang, Naiyan Wang, and Zhaoxiang Zhang.
\newblock Rangedet: In defense of range view for lidar-based 3d object
  detection.
\newblock {\em arXiv preprint arXiv:2103.10039}, 2021.

\bibitem{KITTI2012}
Andreas Geiger, Philip Lenz, and Raquel Urtasun.
\newblock Are we ready for autonomous driving? the kitti vision benchmark
  suite.
\newblock In {\em 2012 IEEE Conference on Computer Vision and Pattern
  Recognition}, pages 3354--3361. IEEE, 2012.

\bibitem{HomoLoss}
Jiaqi Gu, Bojian Wu, Lubin Fan, Jianqiang Huang, Shen Cao, Zhiyu Xiang, and
  Xian-Sheng Hua.
\newblock Homography loss for monocular 3d object detection.
\newblock {\em arXiv preprint arXiv:2204.00754}, 2022.

\bibitem{SA-SSD}
Chenhang He, Hui Zeng, Jianqiang Huang, Xian-Sheng Hua, and Lei Zhang.
\newblock Structure aware single-stage 3d object detection from point cloud.
\newblock In {\em Proceedings of the IEEE/CVF Conference on Computer Vision and
  Pattern Recognition}, pages 11873--11882, 2020.

\bibitem{BEVDet}
Junjie Huang, Guan Huang, Zheng Zhu, and Dalong Du.
\newblock Bevdet: High-performance multi-camera 3d object detection in
  bird-eye-view.
\newblock {\em arXiv preprint arXiv:2112.11790}, 2021.

\bibitem{MonoDTR}
Kuan-Chih Huang, Tsung-Han Wu, Hung-Ting Su, and Winston~H Hsu.
\newblock Monodtr: Monocular 3d object detection with depth-aware transformer.
\newblock {\em arXiv preprint arXiv:2203.10981}, 2022.

\bibitem{Adam}
Diederik~P Kingma and Jimmy Ba.
\newblock Adam: A method for stochastic optimization.
\newblock {\em arXiv preprint arXiv:1412.6980}, 2014.

\bibitem{kumar2022deviant}
Abhinav Kumar, Garrick Brazil, Enrique Corona, Armin Parchami, and Xiaoming
  Liu.
\newblock Deviant: Depth equivariant network for monocular 3d object detection.
\newblock In {\em European Conference on Computer Vision (ECCV)}, 2022.

\bibitem{GrooMeD-NMS}
Abhinav Kumar, Garrick Brazil, and Xiaoming Liu.
\newblock Groomed-nms: Grouped mathematically differentiable nms for monocular
  3d object detection.
\newblock In {\em Proceedings of the IEEE/CVF Conference on Computer Vision and
  Pattern Recognition}, pages 8973--8983, 2021.

\bibitem{lang2019pointpillars}
Alex~H. Lang, Sourabh Vora, Holger Caesar, Lubing Zhou, Jiong Yang, and Oscar
  Beijbom.
\newblock Point{P}illars: Fast encoders for object detection from point clouds.
\newblock In {\em CVPR}, pages 12697--12705, 2019.

\bibitem{RTM3D}
Peixuan Li, Huaici Zhao, Pengfei Liu, and Feidao Cao.
\newblock Rtm3d: Real-time monocular 3d detection from object keypoints for
  autonomous driving.
\newblock {\em arXiv preprint arXiv:2001.03343}, 2020.

\bibitem{BEVDepth}
Yinhao Li, Zheng Ge, Guanyi Yu, Jinrong Yang, Zengran Wang, Yukang Shi,
  Jianjian Sun, and Zeming Li.
\newblock Bevdepth: Acquisition of reliable depth for multi-view 3d object
  detection.
\newblock {\em arXiv preprint arXiv:2206.10092}, 2022.

\bibitem{li2022diversity}
Zhuoling Li, Zhan Qu, Yang Zhou, Jianzhuang Liu, Haoqian Wang, and Lihui Jiang.
\newblock Diversity matters: Fully exploiting depth clues for reliable
  monocular 3d object detection.
\newblock In {\em Proceedings of the IEEE/CVF Conference on Computer Vision and
  Pattern Recognition}, pages 2791--2800, 2022.

\bibitem{BEVFormer}
Zhiqi Li, Wenhai Wang, Hongyang Li, Enze Xie, Chonghao Sima, Tong Lu, Yu Qiao,
  and Jifeng Dai.
\newblock Bevformer: Learning bird’s-eye-view representation from
  multi-camera images via spatiotemporal transformers.
\newblock In {\em Computer Vision--ECCV 2022: 17th European Conference, Tel
  Aviv, Israel, October 23--27, 2022, Proceedings, Part IX}, pages 1--18.
  Springer, 2022.

\bibitem{MonoJSG}
Qing Lian, Peiliang Li, and Xiaozhi Chen.
\newblock Monojsg: Joint semantic and geometric cost volume for monocular 3d
  object detection.
\newblock {\em arXiv preprint arXiv:2203.08563}, 2022.

\bibitem{focal}
Tsung-Yi Lin, Priya Goyal, Ross Girshick, Kaiming He, and Piotr Doll{\'a}r.
\newblock Focal loss for dense object detection.
\newblock In {\em Proceedings of the IEEE international conference on computer
  vision}, pages 2980--2988, 2017.

\bibitem{MonoCon}
Xianpeng Liu, Nan Xue, and Tianfu Wu.
\newblock Learning auxiliary monocular contexts helps monocular 3d object
  detection.
\newblock {\em arXiv preprint arXiv:2112.04628}, 2021.

\bibitem{PETR}
Yingfei Liu, Tiancai Wang, Xiangyu Zhang, and Jian Sun.
\newblock Petr: Position embedding transformation for multi-view 3d object
  detection.
\newblock In {\em Computer Vision--ECCV 2022: 17th European Conference, Tel
  Aviv, Israel, October 23--27, 2022, Proceedings, Part XXVII}, pages 531--548.
  Springer, 2022.

\bibitem{Ground-Aware}
Yuxuan Liu, Yuan Yixuan, and Ming Liu.
\newblock Ground-aware monocular 3d object detection for autonomous driving.
\newblock {\em IEEE Robotics and Automation Letters}, 6(2):919--926, 2021.

\bibitem{AutoShape}
Zongdai Liu, Dingfu Zhou, Feixiang Lu, Jin Fang, and Liangjun Zhang.
\newblock Autoshape: Real-time shape-aware monocular 3d object detection.
\newblock In {\em Proceedings of the IEEE/CVF International Conference on
  Computer Vision}, pages 15641--15650, 2021.

\bibitem{GUPNet}
Yan Lu, Xinzhu Ma, Lei Yang, Tianzhu Zhang, Yating Liu, Qi Chu, Junjie Yan, and
  Wanli Ouyang.
\newblock Geometry uncertainty projection network for monocular 3d object
  detection.
\newblock In {\em Proceedings of the IEEE/CVF International Conference on
  Computer Vision}, pages 3111--3121, 2021.

\bibitem{PatchNet}
Xinzhu Ma, Shinan Liu, Zhiyi Xia, Hongwen Zhang, Xingyu Zeng, and Wanli Ouyang.
\newblock Rethinking pseudo-lidar representation.
\newblock {\em arXiv preprint arXiv:2008.04582}, 2020.

\bibitem{AM3D}
Xinzhu Ma, Zhihui Wang, Haojie Li, Pengbo Zhang, Wanli Ouyang, and Xin Fan.
\newblock Accurate monocular 3d object detection via color-embedded 3d
  reconstruction for autonomous driving.
\newblock In {\em Proceedings of the IEEE International Conference on Computer
  Vision}, pages 6851--6860, 2019.

\bibitem{Monodle}
Xinzhu Ma, Yinmin Zhang, Dan Xu, Dongzhan Zhou, Shuai Yi, Haojie Li, and Wanli
  Ouyang.
\newblock Delving into localization errors for monocular 3d object detection.
\newblock In {\em Proceedings of the IEEE/CVF Conference on Computer Vision and
  Pattern Recognition}, pages 4721--4730, 2021.

\bibitem{RoI10D}
Fabian Manhardt, Wadim Kehl, and Adrien Gaidon.
\newblock Roi-10d: Monocular lifting of 2d detection to 6d pose and metric
  shape.
\newblock In {\em Proceedings of the IEEE Conference on Computer Vision and
  Pattern Recognition}, pages 2069--2078, 2019.

\bibitem{occupancyNet}
Lars Mescheder, Michael Oechsle, Michael Niemeyer, Sebastian Nowozin, and
  Andreas Geiger.
\newblock Occupancy networks: Learning 3d reconstruction in function space.
\newblock In {\em Proceedings of the IEEE/CVF conference on computer vision and
  pattern recognition}, pages 4460--4470, 2019.

\bibitem{nerf}
Ben Mildenhall, Pratul~P Srinivasan, Matthew Tancik, Jonathan~T Barron, Ravi
  Ramamoorthi, and Ren Ng.
\newblock Nerf: Representing scenes as neural radiance fields for view
  synthesis.
\newblock {\em Communications of the ACM}, 65(1):99--106, 2021.

\bibitem{Deep3DBox}
Arsalan Mousavian, Dragomir Anguelov, John Flynn, and Jana Kosecka.
\newblock 3d bounding box estimation using deep learning and geometry.
\newblock In {\em Proceedings of the IEEE Conference on Computer Vision and
  Pattern Recognition}, pages 7074--7082, 2017.

\bibitem{noh2021hvpr}
Jongyoun Noh, Sanghoon Lee, and Bumsub Ham.
\newblock Hvpr: Hybrid voxel-point representation for single-stage 3d object
  detection.
\newblock In {\em Proceedings of the IEEE/CVF Conference on Computer Vision and
  Pattern Recognition}, pages 14605--14614, 2021.

\bibitem{DD3D}
Dennis Park, Rares Ambrus, Vitor Guizilini, Jie Li, and Adrien Gaidon.
\newblock Is pseudo-lidar needed for monocular 3d object detection?
\newblock In {\em Proceedings of the IEEE/CVF International Conference on
  Computer Vision}, pages 3142--3152, 2021.

\bibitem{Pytorch}
Adam Paszke, Sam Gross, Francisco Massa, Adam Lerer, James Bradbury, Gregory
  Chanan, Trevor Killeen, Zeming Lin, Natalia Gimelshein, Luca Antiga, et~al.
\newblock Pytorch: An imperative style, high-performance deep learning library.
\newblock In {\em Advances in neural information processing systems}, pages
  8026--8037, 2019.

\bibitem{peng2022did}
Liang Peng, Xiaopei Wu, Zheng Yang, Haifeng Liu, and Deng Cai.
\newblock Did-m3d: Decoupling instance depth for monocular 3d object detection.
\newblock In {\em European Conference on Computer Vision}, 2022.

\bibitem{qi2018frustum}
Charles~R Qi, Wei Liu, Chenxia Wu, Hao Su, and Leonidas~J Guibas.
\newblock Frustum pointnets for 3d object detection from rgb-d data.
\newblock In {\em Proceedings of the IEEE conference on computer vision and
  pattern recognition}, pages 918--927, 2018.

\bibitem{PointNet}
Charles~R Qi, Hao Su, Kaichun Mo, and Leonidas~J Guibas.
\newblock Pointnet: Deep learning on point sets for 3d classification and
  segmentation.
\newblock In {\em Proceedings of the IEEE conference on computer vision and
  pattern recognition}, pages 652--660, 2017.

\bibitem{PointNet++}
Charles~R Qi, Li Yi, Hao Su, and Leonidas~J Guibas.
\newblock Pointnet++: Deep hierarchical feature learning on point sets in a
  metric space.
\newblock {\em arXiv preprint arXiv:1706.02413}, 2017.

\bibitem{CaDDN}
Cody Reading, Ali Harakeh, Julia Chae, and Steven~L Waslander.
\newblock Categorical depth distribution network for monocular 3d object
  detection.
\newblock In {\em Proceedings of the IEEE/CVF Conference on Computer Vision and
  Pattern Recognition}, pages 8555--8564, 2021.

\bibitem{OFTNet}
Thomas Roddick, Alex Kendall, and Roberto Cipolla.
\newblock Orthographic feature transform for monocular 3d object detection.
\newblock {\em arXiv preprint arXiv:1811.08188}, 2018.

\bibitem{shi2020pv}
Shaoshuai Shi, Chaoxu Guo, Li Jiang, Zhe Wang, Jianping Shi, Xiaogang Wang, and
  Hongsheng Li.
\newblock {PV-RCNN}: Point-voxel feature set abstraction for {3D} object
  detection.
\newblock In {\em CVPR}, pages 10529--10538, 2020.

\bibitem{PVRCNN}
Shaoshuai Shi, Chaoxu Guo, Li Jiang, Zhe Wang, Jianping Shi, Xiaogang Wang, and
  Hongsheng Li.
\newblock Pv-rcnn: Point-voxel feature set abstraction for 3d object detection.
\newblock In {\em Proceedings of the IEEE/CVF Conference on Computer Vision and
  Pattern Recognition}, pages 10529--10538, 2020.

\bibitem{Pointrcnn}
Shaoshuai Shi, Xiaogang Wang, and Hongsheng Li.
\newblock Pointrcnn: 3d object proposal generation and detection from point
  cloud.
\newblock In {\em Proceedings of the IEEE/CVF Conference on Computer Vision and
  Pattern Recognition}, pages 770--779, 2019.

\bibitem{PART}
Shaoshuai Shi, Zhe Wang, Jianping Shi, Xiaogang Wang, and Hongsheng Li.
\newblock From points to parts: 3d object detection from point cloud with
  part-aware and part-aggregation network.
\newblock {\em arXiv preprint arXiv:1907.03670}, 2019.

\bibitem{Point-gnn}
Weijing Shi and Raj Rajkumar.
\newblock Point-gnn: Graph neural network for 3d object detection in a point
  cloud.
\newblock In {\em Proceedings of the IEEE/CVF Conference on Computer Vision and
  Pattern Recognition}, pages 1711--1719, 2020.

\bibitem{MonoRCNN}
Xuepeng Shi, Qi Ye, Xiaozhi Chen, Chuangrong Chen, Zhixiang Chen, and Tae-Kyun
  Kim.
\newblock Geometry-based distance decomposition for monocular 3d object
  detection.
\newblock In {\em Proceedings of the IEEE/CVF International Conference on
  Computer Vision}, pages 15172--15181, 2021.

\bibitem{AP40}
Andrea Simonelli, Samuel~Rota Bulo, Lorenzo Porzi, Manuel L{\'o}pez-Antequera,
  and Peter Kontschieder.
\newblock Disentangling monocular 3d object detection.
\newblock In {\em Proceedings of the IEEE International Conference on Computer
  Vision}, pages 1991--1999, 2019.

\bibitem{One}
Leslie~N Smith.
\newblock A disciplined approach to neural network hyper-parameters: Part
  1--learning rate, batch size, momentum, and weight decay.
\newblock {\em arXiv preprint arXiv:1803.09820}, 2018.

\bibitem{waymo}
Pei Sun, Henrik Kretzschmar, Xerxes Dotiwalla, Aurelien Chouard, Vijaysai
  Patnaik, Paul Tsui, James Guo, Yin Zhou, Yuning Chai, Benjamin Caine, et~al.
\newblock Scalability in perception for autonomous driving: Waymo open dataset.
\newblock In {\em Proceedings of the IEEE/CVF Conference on Computer Vision and
  Pattern Recognition}, pages 2446--2454, 2020.

\bibitem{DDMP-3D}
Li Wang, Liang Du, Xiaoqing Ye, Yanwei Fu, Guodong Guo, Xiangyang Xue, Jianfeng
  Feng, and Li Zhang.
\newblock Depth-conditioned dynamic message propagation for monocular 3d object
  detection.
\newblock In {\em Proceedings of the IEEE/CVF Conference on Computer Vision and
  Pattern Recognition}, pages 454--463, 2021.

\bibitem{PCT}
Li Wang, Li Zhang, Yi Zhu, Zhi Zhang, Tong He, Mu Li, and Xiangyang Xue.
\newblock Progressive coordinate transforms for monocular 3d object detection.
\newblock {\em Advances in Neural Information Processing Systems}, 34, 2021.

\bibitem{DFM}
Tai Wang, Jiangmiao Pang, and Dahua Lin.
\newblock Monocular 3d object detection with depth from motion.
\newblock {\em arXiv preprint arXiv:2207.12988}, 2022.

\bibitem{PseudoLidar}
Yan Wang, Wei-Lun Chao, Divyansh Garg, Bharath Hariharan, Mark Campbell, and
  Kilian~Q Weinberger.
\newblock Pseudo-lidar from visual depth estimation: Bridging the gap in 3d
  object detection for autonomous driving.
\newblock In {\em Proceedings of the IEEE Conference on Computer Vision and
  Pattern Recognition}, pages 8445--8453, 2019.

\bibitem{wu2022transformation}
Hai Wu, Chenglu Wen, Wei Li, Xin Li, Ruigang Yang, and Cheng Wang.
\newblock Transformation-equivariant 3d object detection for autonomous
  driving.
\newblock {\em arXiv preprint arXiv:2211.11962}, 2022.

\bibitem{SFD}
Xiaopei Wu, Liang Peng, Honghui Yang, Liang Xie, Chenxi Huang, Chengqi Deng,
  Haifeng Liu, and Deng Cai.
\newblock Sparse fuse dense: Towards high quality 3d detection with depth
  completion.
\newblock In {\em Proceedings of the IEEE/CVF Conference on Computer Vision and
  Pattern Recognition}, pages 5418--5427, 2022.

\bibitem{yan2018second}
Yan Yan, Yuxing Mao, and Bo Li.
\newblock {SECOND}: Sparsely embedded convolutional detection.
\newblock {\em Sensors}, 18(10):3337, 2018.

\bibitem{yang20203dssd}
Zetong Yang, Yanan Sun, Shu Liu, and Jiaya Jia.
\newblock {3DSSD}: Point-based {3D} single stage object detector.
\newblock In {\em CVPR}, pages 11040--11048, 2020.

\bibitem{yang2019std}
Zetong Yang, Yanan Sun, Shu Liu, Xiaoyong Shen, and Jiaya Jia.
\newblock {STD}: Sparse-to-dense {3D} object detector for point cloud.
\newblock In {\em ICCV}, pages 1951--1960, 2019.

\bibitem{Yin_2021_CVPR}
Tianwei Yin, Xingyi Zhou, and Philipp Krahenbuhl.
\newblock Center-based 3d object detection and tracking.
\newblock In {\em Proceedings of the IEEE/CVF Conference on Computer Vision and
  Pattern Recognition (CVPR)}, pages 11784--11793, June 2021.

\bibitem{DLA34}
Fisher Yu, Dequan Wang, Evan Shelhamer, and Trevor Darrell.
\newblock Deep layer aggregation.
\newblock In {\em Proceedings of the IEEE conference on computer vision and
  pattern recognition}, pages 2403--2412, 2018.

\bibitem{nerf++}
Kai Zhang, Gernot Riegler, Noah Snavely, and Vladlen Koltun.
\newblock Nerf++: Analyzing and improving neural radiance fields.
\newblock {\em arXiv preprint arXiv:2010.07492}, 2020.

\bibitem{MonoFlex}
Yunpeng Zhang, Jiwen Lu, and Jie Zhou.
\newblock Objects are different: Flexible monocular 3d object detection.
\newblock In {\em Proceedings of the IEEE/CVF Conference on Computer Vision and
  Pattern Recognition}, pages 3289--3298, 2021.

\bibitem{zheng2021se}
Wu Zheng, Weiliang Tang, Li Jiang, and Chi-Wing Fu.
\newblock Se-ssd: Self-ensembling single-stage object detector from point
  cloud.
\newblock In {\em Proceedings of the IEEE/CVF Conference on Computer Vision and
  Pattern Recognition}, pages 14494--14503, 2021.

\bibitem{MonoEF}
Yunsong Zhou, Yuan He, Hongzi Zhu, Cheng Wang, Hongyang Li, and Qinhong Jiang.
\newblock Monocular 3d object detection: An extrinsic parameter free approach.
\newblock In {\em Proceedings of the IEEE/CVF Conference on Computer Vision and
  Pattern Recognition}, pages 7556--7566, 2021.

\bibitem{zhou2018voxelnet}
Yin Zhou and Oncel Tuzel.
\newblock Voxelnet: End-to-end learning for point cloud based 3d object
  detection.
\newblock In {\em Proceedings of the IEEE conference on computer vision and
  pattern recognition}, pages 4490--4499, 2018.

\end{thebibliography}
}

\end{document}